\documentclass[10pt,twoside]{article}

\usepackage{times}
\usepackage[utf8]{inputenc}
\usepackage[T1]{fontenc}
\usepackage{graphicx}
\usepackage{hyperref}
\usepackage{comment}

\usepackage{siunitx}
\usepackage{taln2022}
\usepackage[french]{babel} 

\usepackage{tikz}
\usetikzlibrary{matrix,chains,positioning,decorations.pathreplacing,arrows}
\usetikzlibrary{positioning,calc}
\usetikzlibrary{decorations.pathmorphing} 
\usetikzlibrary{fit}					
\usetikzlibrary{backgrounds}	

\title{Vers la compréhension automatique de la parole bout-en-bout à moindre effort}

\author{Marco Naguib\quad François Portet\quad Marco Dinarelli\\
  {\small
    Univ. Grenoble Alpes, CNRS, Grenoble INP, LIG, 38000 Grenoble, France \\
    \texttt{
      marco.naguib@hotmail.com, (francois.portet|marco.dinarelli)@univ-grenoble-alpes.fr} \\ 
}}

\begin{document}
\maketitle

\resume{
Les approches de compréhension automatique de la parole ont récemment bénéficié de l'apport de modèles préappris par autosupervision sur de gros corpus de parole. Pour le français, le projet \emph{LeBenchmark} a rendu disponibles de tels modèles et a permis des évolutions impressionnantes sur plusieurs tâches dont la compréhension automatique de la parole. Ces avancées ont un coût non négligeable en ce qui concerne le temps de calcul et la consommation énergétique. 
Dans cet article, nous comparons plusieurs stratégies d'apprentissage visant à réduire le coût énergétique tout en conservant des performances compétitives. Les expériences sont effectuées sur le corpus MEDIA, et montrent qu'il est possible de réduire significativement le coût d'apprentissage tout en conservant des performances à l'état de l'art. 
}

\abstract{Towards automatic end-to-end speech understanding with less effort}{
  Recent advances in spoken language understanding benefited from Self-Supervised models trained on large speech corpora. For French, the LeBenchmark project has made such models available and has led to impressive progress on several tasks including spoken language understanding. These advances have a non-negligible cost in terms of computation time and energy consumption. 
In this paper, we compare several learning strategies aiming at reducing such cost while keeping competitive performances. The experiments are performed on the MEDIA corpus, and show that it is possible to reduce the learning cost while maintaining state-of-the-art performances. 
}

\begin{small}
\motsClefs
  {compréhension de la parole, apprentissage autosupervisé,  apprentissage par transfert}
  {Spoken Language Understanding, Self-Supervised Learning, Transfer Learning}
\end{small}



\section{Introduction}

La compréhension automatique de la parole (SLU de \emph{Spoken Language Understanding}) vise à extraire une représentation sémantique à partir d'un signal audio contenant un énoncé en langage naturel \cite{DeMori1997:SDBook}. Les approches classiques utilisées pour extraire la sémantique de la parole ont consisté à mettre en cascade un module de reconnaissance automatique de la parole avec un système de compréhension du langage naturel \cite{Raymond2006:FST-SEM,dinarelli09:emnlp,dinarelli09:Interspeech,Hahn.etAL-SLUJournal-2010,Dinarelli2010.PhDThesis,caubriere:hal-02465899,ghannay:hal-03372494}. Les réseaux de neurones ont permis l'avancement des systèmes bout-en-bout pour la SLU \cite{DBLP:journals/corr/abs-1802-08395,desot:hal-02464393,lugosch2019speech,DBLP:journals/corr/abs-1906-07601,dinarelli2020data,pelloin:hal-03128163}, qui sont préférés aux systèmes en cascade, notamment pour leur capacité à réduire l'effet d'erreur en cascade et à exploiter des composantes acoustiques pour déduire certaines informations sémantiques \cite{desot:hal-02464393}. \\
Bien que des approches ont proposé un apprentissage de bout en bout du modèle \cite{8268987,desot:hal-02464393,palogiannidi2020endtoend}, de nombreux travaux ont appliqué un apprentissage graduel du modèle sur des tâches de plus en plus spécifiques. 
Partant de l'hypothèse qu'un modèle de SLU doit nécessairement apprendre une représentation de la parole, \cite{lugosch2019speech} et \cite{dinarelli2020data} proposent une approche où le modèle est progressivement entraîné à reconnaître la transcription puis à en extraire la sémantique. On peut également citer \cite{serdyuk2018endtoend} et \cite{radfar2020endtoend} qui apprennent, en première étape, un classificateur de domaine auquel se rapporte l'intention, avant d'optimiser le modèle pour classifier les intentions et les attributs sémantiques d'un énoncé. 
Un changement récent dans les approches de SLU de bout en bout est l'utilisation de modèles appris par \emph{auto-supervision} (SSL de \emph{Self-Supervised Learning}) sur de très gros corpus de parole, tels que \emph{wav2vec} ou \emph{HuBERT} \cite{DBLP:journals/corr/abs-1904-05862,NEURIPS2020_92d1e1eb,DBLP:journals/corr/abs-2106-07447}. Dans \cite{Lai2020}, un modèle wav2vec préappris est utilisé comme encodeur de la parole tandis que le benchmark SUPERB \cite{superb} propose une tâche de \emph{slot-filling} et de classification d'intention parmi les tâches d'évaluation des modèles de la parole préappris. 
En 2021, de tels modèles ont été mis à disposition de la communauté française \cite{evain:hal-03317730,evain:hal-03407172}, permettant une amélioration impressionnante des performances sur des tâches telles que la SLU. \\
Si l'on peut se féliciter que les avancés de la recherche ont permis d'améliorer les performances obtenues sur les tâches visées, ces avancées ont un coût non négligeable en ce qui concerne le temps de calcul et la consommation énergétique \cite{parcollet:hal-03190119}.
Même des modèles SSL monolingues \cite{evain:hal-03317730,evain:hal-03407172} 
demandent près de deux semaines d'apprentissage sur $64$ GPUs. On peut arguer que ce coût reste contenu par le fait que ces modèles sont entraînés une fois et utilisés ensuite pour beaucoup d'applications différentes.
Cependant, ces modèles ne constituent souvent qu'un encodeur et doivent être adaptés sur les tâches en aval pour améliorer les performances des systèmes. Cette pratique multiplie davantage les phases d'apprentissage et conduit par conséquent à des consommations de ressources importantes. \\
Dans cet article nous nous intéressons à réduire le coût nécessaire pour obtenir des performances compétitives sur des tâches de SLU en utilisant les modèles SSL déployés par \cite{evain:hal-03317730,evain:hal-03407172}.
Dans cet article, nous proposons une étude visant à trouver un meilleur compromis entre les performances et le coût énergétique.
Pour cela, nous étudions des stratégies d'apprentissage différentes de celles utilisées par \cite{evain:hal-03317730,evain:hal-03407172} que nous couplons à une stratégie d'apprentissage par transfert avec des modèles appris pour d'autres tâches \cite{Lefevre2012}, et à une phase d'affinage d'un modèle SSL français directement sur la tâche SLU, au lieu de la tâche ASR comme proposé dans \cite{evain:hal-03407172}.
Bien que cette dernière soit relativement coûteuse par rapport à l'apprentissage des modèles en aval pour la SLU, elle reste moins lourde que les approches similaires proposées récemment pour la même tâche \cite{pelloin:hal-03128163,ghannay:hal-03372494}, tout en permettant d'obtenir des performances comparables.

\section{Compréhension automatique de la parole et modèles SSL}

\begin{figure}[!ht]
\center
\scalebox{0.5}{
\begin{tikzpicture}[scale = 1.0]
	\begin{scope}[local bounding box=net]
	
	\node [scale=1.5] (EL) at (-2.0,4.7) {Encodeur};
	\node[draw, thick, rectangle, rounded corners=1pt, scale=1.2,
	minimum width=10cm, minimum height=3cm, text=blue] (EB) at (4.0,2.5) {};
	\node (h0) at (-1.5,1.5) {$h_0$};
	\node (i1) at (0,0.0) {$x_1$};
	\node (i2) at (2.0,0.0) {$x_2$};
	\node (i3) at (4.0,0.0) {$x_3$};
	\node (dots) at (6.0,0.0) {$\dots$};
	\node (iN) at (8.0,0.0) {$x_N$};
	
	\node[draw, rectangle, rounded corners=1pt, scale=1.2] (e1) at (0,1.5) {LSTM$_e$};
	\node[draw, rectangle, rounded corners=1pt, scale=1.2] (e2) at (2.0,1.5) {LSTM$_e$};
	\node[draw, rectangle, rounded corners=1pt, scale=1.2] (e3) at (4.0,1.5) {LSTM$_e$};
	\node[draw, rectangle, rounded corners=1pt, scale=1.2] (eN) at (8.0,1.5) {LSTM$_e$};
	
	\draw[->] (h0) -- (e1.west);
    \draw[->] (i1) -- (e1);
    \draw[->] (i2) -- (e2);
    \draw[->] (i3) -- (e3);
    \draw[->] (iN) -- (eN);
	
	\node[draw, rectangle, rounded corners=1pt, scale=1.2] (h1) at (0,3.0) {$h_1^e$};
	\node[draw, rectangle, rounded corners=1pt, scale=1.2] (h2) at (2.0,3.0) {$h_2^e$};
	\node[draw, rectangle, rounded corners=1pt, scale=1.2] (h3) at (4.0,3.0) {$h_3^e$};
	\node[draw, rectangle, rounded corners=1pt, scale=1.2] (hN) at (8.0,3.0) {$h_N^e$};
    
    \draw[->] (e1) -- (h1);
    \draw[->] (h1) -- (e2.west);
    \draw[->] (e2) -- (h2);
    \draw[->] (h2) -- (e3.west);
    \draw[->] (e3) -- (h3);
    \node (dots) at (6.0,2.0) {$\dots$};
    \draw[->] (eN) -- (hN);

    \node [teal, scale=1.5] (DL) at (-6.0,11.0) {Decodeur};
	\node[teal, thick, draw, rectangle, rounded corners=1pt, scale=1.2,
	minimum width=11cm, minimum height=4.2cm, text=blue] (DB) at (0.6,8.0) {};
    \node[draw, rectangle, rounded corners=1pt, scale=1.2, text=teal] (dh) at (6.0,8.5) {$h_4^d$};
    \node[draw, rectangle, rounded corners=1pt, scale=1.2, text=teal] (d4) at (6.0,7.0) {LSTM$_d$};
    \node[draw, rectangle, rounded corners=1pt, scale=1.2, text=teal] (s3) at (4.0,7.0) {$[s_3 ; h_3^d]$};
    \node[draw, rectangle, rounded corners=1pt, scale=1.2, text=teal] (ax) at (2.0,6.0) {Att($x$)};
    \node[draw, rectangle, rounded corners=1pt, scale=1.2, text=teal] (ay) at (2.0,8.0) {Att($y$)};
    
    \node[draw, rectangle, rounded corners=1pt, scale=1.2, text=teal] (y1) at (-5.0,8.0) {$E(y_1)$};
    \node[draw, rectangle, rounded corners=1pt, scale=1.2, text=teal] (y2) at (-3.0,8.0) {$E(y_2)$};
    \node[draw, rectangle, rounded corners=1pt, scale=1.2, text=teal] (y3) at (-1.0,8.0) {$E(y_3)$};

    \draw[teal, ->] (dh) to[bend right=30] (ax);
    \draw[teal, ->] (h1) to[bend left=20] (ax);
    \draw[teal, ->] (h2) to[bend left=20] (ax);
    \draw[teal, ->] (h3) to[bend left=20] (ax);
    \node [text=teal] (dots) at (5.0,4.0) {$\dots$};
    \draw[teal, ->] (hN) to[bend right=20] (ax);

    \draw[teal, ->] (s3) -- (d4);
    \draw[teal, ->] (d4) -- (dh);
    \draw[teal, ->] (dh) to[bend right=10] (ay);
    \draw[teal, ->] (y1) to[bend left=30] (ay);
    \draw[teal, ->] (y2) to[bend right=30] (ay);
    \draw[teal, ->] (y3) -- (ay);
    
    \node[draw, rectangle, rounded corners=1pt, scale=1.2, text=blue] (s) at (6.0,10.0) {$s_4$};
    \node [blue] (y) at (6.0,11.5) {$y_4$};
    \draw[blue, ->] (ax) to[bend left=20] (s);
    \draw[blue, ->] (ay) to[bend left=30] (s);
    \draw[blue, ->] (s) -- (y);

    \end{scope}

\end{tikzpicture}
}
    \caption{\emph{Schéma de notre architecture neuronale pour la SLU}}\label{fig:E2ESLU}
\end{figure}
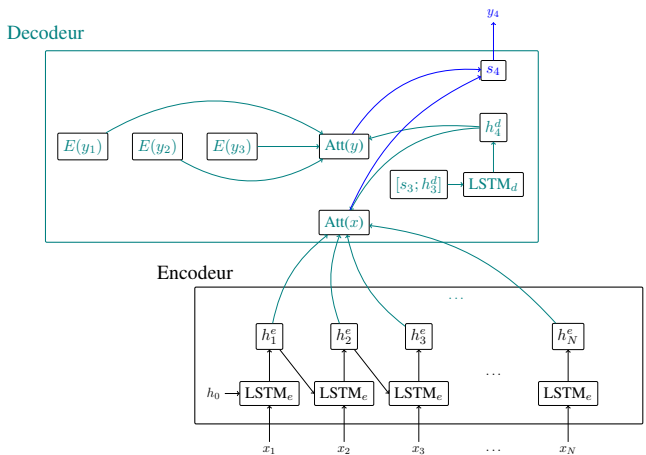

Dans cet article, nous exploitons des modèles préappris de manière autosupervisée et cherchons à tirer parti de ceux-ci de la manière la plus économe pour une tâche de SLU.  
Nous utiliserons les modèles \textit{LeBenchmark} \cite{evain:hal-03407172}. 
Ces modèles ont été évalués sur quatre tâches dont la SLU. 
Sur cette dernière, le modèle \emph{w2v2-fr-7k} a montré les meilleurs résultats, c'est pourquoi nous l'utilisons dans nos expériences. \\
\textbf{Les modèles pour la SLU} que nous utilisons sont les mêmes que ceux utilisés dans \cite{evain:hal-03407172}\footnote{Nous avons téléchargé les modèles disponibles sur \url{https://huggingface.co/LeBenchmark} et les systèmes disponibles sur \url{https://github.com/LeBenchmark/NeurIPS2021}}. Il s'agit de modèles \emph{encodeur-décodeur} basés sur les LSTM et le mécanisme d'attention \cite{Hochreiter-1997-LSTM,DBLP:journals/corr/BahdanauCB14}.
L'encodeur a une structure pyramidale similaire à \cite{DBLP:journals/corr/ChanJLV15} tandis que le décodeur intègre deux mécanismes d'attention, un pour atteindre les états cachés de l'encodeur, l'autre pour atteindre toutes les prédictions précédentes, comme le module de \emph{self-attention} des Transformers \cite{NIPS2017_3f5ee243}.
Un schéma de cette architecture neuronale est montrée dans la figure~\ref{fig:E2ESLU}, où nous différencions les éléments de l'encodeur et du décodeur avec les exposants $^e$ et $^d$ respectivement. $E(y_i)$ indique le plongement de l'étiquette $y_i$, les deux mécanismes d'attention sont indiqués respectivement avec $Att(x)$ et $Att(y)$.
Les modèles sont appris en minimisant la fonction de coût CTC \cite{Graves:2006:CTC:1143844.1143891}. Le choix du meilleur modèle est fait sur la base du taux d'erreur sur les données de développement de la tâche visée, en considérant uniquement les tours de parole utilisateur (cf section~\ref{subsec:data}).
Dans ce travail, nous utilisons les modèles SSL pour extraire les \emph{features} qui alimentent les modèles SLU comme alternative aux paramètres classiques (MFCC, spectrogrammes, etc.). \\
Les modèles décrits dans \cite{evain:hal-03407172} demandent au total trois étapes d'apprentissage, stratégie indiquée avec \emph{3 steps} dans les tableaux, chacune utilisant le modèle de l'étape précédente pour l'initialisation des paramètres du modèle~: {\bf 1)} apprentissage de l'encodeur sur la transcription~; {\bf 2)} apprentissage de l'encodeur sur la SLU~; {\bf 3)} apprentissage du modèle final, encodeur et décodeur, sur la SLU.
Bien que cette stratégie soit la plus efficace, elle demande un coût d'apprentissage important, d'autant plus que les modèles les plus performants dans \cite{evain:hal-03407172} demandent une étape additionnelle et coûteuse d'adaptation du modèle SSL sur la tâche finale. \\
Les performances des modèles SLU avec entrée produite par un modèle SSL sont très élevées, intuitivement donc des performances similaires, ou légèrement inférieures, peuvent être obtenues avec un coût d'apprentissage moindre, permettant d'économiser des ressources.
Nous souhaitons avec nos analyses trouver un meilleur compromis entre effort d'apprentissage et performance finale du modèle. \\ 
Afin de valider cette hypothèse nous essayons alors deux stratégies d'apprentissage différentes par rapport à \cite{evain:hal-03407172} (\textbf{Stg appr.} dans les tableaux)~: \textit{2 steps}, nous effectuons uniquement les étapes 2 et 3 de la stratégie \emph{3 steps}~; 
\textit{1 step}, nous entraînons directement le modèle final sur la SLU.
Nous notons que cette dernière stratégie correspond à un apprentissage \textit{réellement} de bout en bout d'un modèle SLU.


\section{Évaluation}
\label{sec:eval}

\subsection{Évaluation du coût d'apprentissage}
\label{subsec:cost}

Les résultats quantitatifs de la SLU sont évalués avec le taux d'erreur sur les concepts (CER -- \emph{Concept Error Rate}).
Nous évaluons le coût computationnel de nos modèles en mesurant~: le temps d'apprentissage, la consommation électrique en \emph{kWh} et sa conversion en grammes de CO$_{2}$ (gCO$_2$ dans les tableaux). Ces 2 dernières valeurs sont obtenues grâce au logiciel \emph{codecarbon}\footnote{\url{https://codecarbon.io}}. Puisque celui-ci surestime le coefficient de conversion de kWh vers grammes de CO$_{2}$, comme dans \cite{parcollet:hal-03190119} nous utilisons le coefficient officiel de $51$ grammes/kWh.\footnote{Disponible sur \url{https://www.eea.europa.eu/data-and-maps/indicators/overview-of-the-electricity-production-3/assessment}.}
Afin d'avoir un cadre plus complet nous montrons également le coût en \emph{kWh} par point de CER gagné (indiqué avec \textbf{kWh/p} dans les tableaux). Pour obtenir cette valeur, nous présumons qu'un modèle étalon $\mathcal{M}_e$ moins coûteux obtient des résultats inférieur à un modèle comparé $\mathcal{M}_c$. En indiquant alors avec $\text{kWh}(\mathcal{M}_i)$ et $\text{CER}(\mathcal{M}_i)$ respectivement la consommation énergétique et le CER du modèle $\mathcal{M}_i$, la valeur \emph{kWh/p} est obtenue par $\frac{\text{kWh}(\mathcal{M}_c) - \text{kWh}(\mathcal{M}_e)}{\text{CER}(\mathcal{M}_e) - \text{CER}(\mathcal{M}_c)}$, avec la contrainte $\text{kWh}(\mathcal{M}_c) \ge \text{kWh}(\mathcal{M}_e)$.
Puisque $\mathcal{M}_e$ est moins coûteux, le numérateur est positif. En présumant $\mathcal{M}_e$ moins performant, le dénominateur est aussi positif puisque le CER est plus faible pour un modèle plus performant. Dans les tableaux suivants, le \emph{kWh/p} est donné par rapport au modèle le moins coûteux entraîné avec les mêmes \emph{features} en entrée. Ce modèle est indiqué avec $\mathcal{M}_e$ dans les tableaux. Quand un modèle $\mathcal{M}_c$ est plus coûteux et moins performant que le modèle étalon $\mathcal{M}_e$, nous utilisons par convention la valeur $\infty$, indiquant qu'une stratégie d'apprentissage plus coûteuse ne donnerait aucun gain en performances. \\
L'interprétation de la valeur \emph{kWh/p} doit être faite gardant en tête l'hypothèse selon laquelle un modèle moins coûteux est aussi moins performant. Alors, pour des features en entrée données, \emph{kWh/p} mesure le coût additionnel en kWh pour améliorer hypothétiquement les résultats d'un point. Ce coût pourrait être dû à l'utilisation d'un modèle plus grand et/ou plus de données d'apprentissage.

\subsection{Données}
\label{subsec:data}

\begin{table}[t]
\begin{center}
\scalebox{0.6}{
    \begin{tabular}{|l|rr|rr|rr|}
      \hline
      & \multicolumn{2}{|c|}{Train} & \multicolumn{2}{|c|}{Dev} & \multicolumn{2}{|c|}{Test}\\
      \hline
      Durée totale audio     &\multicolumn{2}{|c|}{27.92h} &\multicolumn{2}{|c|}{6.27h}
      &\multicolumn{2}{|c|}{5.79h} \\
      \multicolumn{1}{|r|}{dont utilisateur}     &\multicolumn{2}{|c|}{8.49h} &\multicolumn{2}{|c|}{2.02h}
      &\multicolumn{2}{|c|}{1.79h} \\
      \# phrases     &\multicolumn{2}{|c|}{13~452} &\multicolumn{2}{|c|}{3~067}&\multicolumn{2}{|c|}{2~886} \\
      \multicolumn{1}{|r|}{ dont utilisateur}     &\multicolumn{2}{|c|}{6~495} &\multicolumn{2}{|c|}{1~485}&\multicolumn{2}{|c|}{1~391} \\
      \hline
      \hline
      & Mots & Étiquettes &  Mots & Étiquettes & Mots & Étiquettes \\
      \hline
      \# tokens	& 212~301 & 24~065 & 47~101 & 5~410 & 44~850 & 4~956 \\
      \multicolumn{1}{|r|}{dont utilisateur}	& 45~710 & 17~064 & 10~465 & 3~814 & 9~898 & 3~451 \\
      dictionnaire		&  2~292 &     37 &    1~339 &    33 &  1~168 &     29 \\
      OOV\%	& -     & -     &  0.30  & 0.0  & 0.34 & 0.0 \\
      \hline
    \end{tabular}
}
    \caption{Statistiques du corpus PortMEDIA}
  \label{tab:PortMEDIAStats}
\end{center}
\end{table}

\begin{table}[t]
\begin{center}
\scalebox{0.6}{
    \begin{tabular}{|l|rr|rr|rr|}
      \hline
      & \multicolumn{2}{|c|}{Train} & \multicolumn{2}{|c|}{Dev} & \multicolumn{2}{|c|}{Test}\\
      \hline
      Durée totale audio     &\multicolumn{2}{|c|}{41.27h} &\multicolumn{2}{|c|}{3.60h}
      &\multicolumn{2}{|c|}{11.28h} \\
      \multicolumn{1}{|r|}{dont utilisateur}     &\multicolumn{2}{|c|}{16.58h} &\multicolumn{2}{|c|}{1.63h}
      &\multicolumn{2}{|c|}{4.59h} \\
      \# phrases     &\multicolumn{2}{|c|}{26~966} &\multicolumn{2}{|c|}{2~662}&\multicolumn{2}{|c|}{6~789} \\
      \multicolumn{1}{|r|}{dont utilisateur}     &\multicolumn{2}{|c|}{12~887} &\multicolumn{2}{|c|}{1~259}&\multicolumn{2}{|c|}{3~005} \\
      \hline
      \hline
      & Mots & Étiquettes &  Mots & Étiquettes & Mots & Étiquettes \\
      \hline
      \# tokens	& 286~327 & 57~915 & 28~213 & 6~219 & 76~591 & 15~418 \\
      \multicolumn{1}{|r|}{dont utilisateur}	& 95~881 & 43~832 & 11~049 & 4~816 & 25~921 & 11~632 \\
      dictionnaire		&  2~785 &     71 &    1~032 &    59 &  1~310 &     67 \\
      OOV\%	& -     & -     &  0.0  & 1,69  & 0.0 & 2.99 \\
      \hline
    \end{tabular}
}
    \caption{Statistiques du corpus MEDIA}
  \label{tab:MEDIAStats}
\end{center}
\end{table}

Si le nombre de corpus pour la SLU est important en langue anglaise, il existe un nombre relativement restreint de ce type de corpus pour le français. On peut citer MEDIA \cite{bonneau-maynard-etal-2006-results}, PORTMEDIA \cite{Lefevre2012}, le HIS~\cite{fleury:hal-00799697}, Sweet-Home~\cite{vacher2014} Vocadom~\cite{vocadom}, ou encore Voice-Home-2~\cite{voicehome2}. Nous utiliserons principalement le corpus MEDIA \cite{bonneau-maynard-etal-2006-results}, dans le domaine de la réservation hôtelière, que nous avons largement utilisé dans le passé \cite{Quarteroni.etAl:Interspeech09,Dinarelli2010:sds,2016:arXiv:DinarelliTellier:NewRNN,Dupont.etAl:LDRNN:CICling2017,dinarelli:hal-01553830}. Les statistiques pour les partitions de données d'apprentissage (Train), de développement (Dev) et de test (Test) sont montrées dans le tableau~\ref{tab:MEDIAStats}.
Ce corpus est constitué de $1~250$ dialogues humain-machine acquis avec une approche par Magicien d'Oz, où $250$ utilisateurs ont suivi $5$ scénarios de réservation. Les signaux de parole ont été
transcrits et annotés avec 76 concepts sémantiques. 
Le corpus est composé de $12~908$ énoncés ($41,5\,h$) pour l'entraînement, $1~259$ énoncés ($3,5\,h$) pour le développement et $3~005$ énoncés ($11,3\,h$) pour le test.
Les sessions de dialogue mettant en scène un utilisateur et un magicien, seuls les tours de parole des utilisateurs ont été annotés avec des concepts et peuvent être utilisés pour entraîner les modèles SLU. Dans nos expériences, nous avons constaté cependant qu'en utilisant aussi les tours de parole du magicien d'Oz pour la SLU les résultats s'améliorent. Pour ce faire, nous avons construit automatiquement le format d'annotation SLU pour les tours de parole du magicien en leur associant le concept conventionnel \emph{MachineSemantic}. Pour que le modèle puisse alors distinguer entre les mots qui instancient de vrais concepts dans les tours utilisateur et les mêmes mots dans les tours du magicien, un marqueur d'\emph{orateur} est ajouté dans les signaux audio en entrée.\footnote{Ce marqueur est constitué d'un tenseur de taille 3 ajouté en tête et à la fin du signal original. Le tenseur contient uniquement des valeurs $+5.0$ pour l'utilisateur et  uniquement des valeurs $-5.0$ pour le magicien.} Par contre, l'ensemble des tours de parole (magicien et utilisateur) ont été transcrits manuellement et peuvent donc être utilisés pour entraîner un modèle ASR. Dans le tableau~\ref{tab:MEDIAStats}, nous montrons ainsi à la fois les statistiques sur l'ensemble des données ainsi que les statistiques pour les tours de parole des utilisateurs seulement (indiqué avec \emph{dont utilisateur} dans le tableau).
Dans cet article, la tâche MEDIA est considérée comme tâche cible, c'est-à-dire la tâche sur laquelle nous souhaitons obtenir les meilleurs résultats possibles à moindre coût, en partant éventuellement de ressources déjà disponibles telles que des modèles SSL et des modèles SLU preappris sur d'autres tâches. \\
Afin de tester l'intérêt d'un transfert d'apprentissage et suivant \cite{caubriere:hal-02304597} nous avons également considéré le corpus PORTMEDIA \cite{Lefevre2012} dédié à la réservation de billets pour le Festival d'Avignon 2010. Le corpus a été acquis et annoté en suivant le même paradigme que MEDIA afin de minimiser les différences entre les 2 corpus (hormis le domaine). 
Il est également divisé en trois parties : un ensemble d'entraînement contenant $5~900$ énoncés, une partie développement contenant $1~400$ énoncés, et un ensemble de test contenant $2~800$ énoncés. Le corpus PORTMEDIA a été annoté manuellement avec 36 concepts sémantiques proches de l'ensemble de concepts MEDIA : PORTMEDIA et MEDIA partagent 26 concepts sémantiques communs.
Les statistiques sur les partitions de données d'apprentissage (\emph{Train}), de développement (\emph{Dev}) et de test (\emph{Test}) pour les corpus PortMEDIA sont montrées dans le tableau~\ref{tab:PortMEDIAStats}. Les considérations faites sur le corpus MEDIA concernant la répartition des données en tours de parole des magiciens d'Oz et des utilisateurs valent également pour PortMEDIA.

\subsection{Résultats}
\label{subsec:results}

\begin{table}[th]
\begin{center}
\scalebox{0.8}{
    \begin{tabular}{l|c|c|c|c|c|c}

        \hline
        \multicolumn{7}{c}{Corpus: PortMEDIA, Métrique: taux d'erreur (CER)} \\
        \hline
        \textbf{Stg appr.} & \textbf{Entrée} & \textbf{KWh} (gCO2) & \textbf{kWh/p} & \textbf{T appr.} & \textbf{DEV} & \textbf{TEST} \\
        \hline
        \hline
        \multicolumn{7}{|c|}{\textbf{Features de base}} \\
        \hline
        3 steps   &   spectro & 4,473 (228) & 0,099 & 36h14' &   35.91  &   40.57 \\ 
        2 steps   &   spectro & 2,989 (152) & $\infty$ & 24h14' &   65.80  &   87.32 \\ 
        1 step   &   spectro & 1,708 (87) & $\mathcal{M}_e$ & 15h52' &   59.22  &   68.50 \\
        \hline
        \hline
        3 steps   &   w2v2-fr & 3,983 (203) & 2,235 & 36h22' &   22.17 &   22.51 \\ 
        2 steps   &   w2v2-fr & 2,707 (138) & 1,939 & 24h27' &   21.86  &   23.02 \\ 
        1 step   &   w2v2-fr & 1,815 (93) & $\mathcal{M}_e$ & 18h08' &   25.53  &   23.48 \\
        \hline
        \hline
        \multicolumn{7}{|c|}{\textbf{Features affinés (+100h x4 GPU)}} \\
        \hline
        1 step +1   &   w2v2-fr slu & 1,214 (62) & - & 11h34' &   21.50  &   22.13 \\ 
        \hline
        
    \end{tabular}
}
\end{center}
\caption{Résultats sur le corpus PortMEDIA,  pour tous les détails voir dans le texte.}
\label{tab:PortMEDIA}
\end{table}

\subsubsection{Expériences préliminaires sur \emph{PortMEDIA}}
\label{subsubsec:portmedia}

Afin obtenir des conditions favorables à l'entretient de modèles à moindre coût sur notre tâche cible (MEDIA), nous avons entraîné, en plus des modèles SSL pour le français \cite{evain:hal-03407172}, des modèles SLU sur la tâche PortMEDIA. Ces modèles seront ensuite utilisés pour préinitialiser les modèles pour la tâche MEDIA.
Les résultats sont montrés dans le tableau~\ref{tab:PortMEDIA}.
Pour avoir une vue globale, nous avons entraîné des modèles à la fois avec les \emph{features} de base (spectrogrammes, indiqué avec \emph{spectro}) et avec les features produits par le modèle SSL pour le français \emph{w2v2-fr 7k} \cite{evain:hal-03407172}. 
Comme on peut le voir, les meilleurs résultats sont toujours obtenus avec la stratégie d'apprentissage la plus coûteuse \emph{3 steps}. Cependant, en utilisant les features du modèle \emph{w2v2-fr} la différence entre la stratégie \emph{3 steps} et la stratégie \emph{1 step} est de moins d'un point de CER, alors que cette dernière stratégie est bien moins coûteuse, à la fois en termes de temps d'apprentissage que de consommation énergétique.
Des modèles encore plus performants en termes de CER peuvent être obtenus, et ce à un coût encore inférieur, avec des features extraits à partir du modèle \emph{w2v2-fr} affiné sur la tâche SLU MEDIA, montrés dans la dernière ligne du tableau~\ref{tab:PortMEDIA} (features \emph{w2v2-fr slu}).
Ces résultats sont accompagnés avec \emph{1 step +1} pour prendre en compte l'affinage du modèle sur la tâche MEDIA, ce qui demande 100 heures d'apprentissage sur 4 GPUs (\textbf{+100h x4 GPU} dans les tableaux).
Le coût de cet apprentissage domine le coût d'apprentissage du modèle SLU. 
Comme nous l'avions mentionné cependant, cet affinage est effectué sur la tâche MEDIA et une fois pour toutes. 
Nous avons choisi d'optimiser sur la tâche MEDIA seulement, à la fois pour réduire le coût d'apprentissage total, et parce que le modèle SLU entraîné sur la tâche PortMEDIA est utilisé par la suite pour préinitialiser le modèle pour la tâche MEDIA.
Comme nous l'avions anticipé donc, grâce à des modèles SSL pour le français, il est possible d'obtenir des performances compétitives sur la tâche SLU visée à un coût bien moindre (stratégie \emph{3 steps} vs stratégie \emph{1 step}). Logiquement les résultats sont encore meilleurs si on dispose d'un modèle SSL affiné. \\
Grâce à ces premières expériences, nous disposons de modèles SLU, en plus des modèles SSL, utilisables pour effectuer un apprentissage par transfert sur la tâche MEDIA.
Le transfert est effectué en préinitialisant les modèles appris sur MEDIA avec un modèle appris sur PortMEDIA.
Dans un contexte 
réel, il est souhaitable que ces ressources existent à l'avance, et qu'elles soient re-utilisées en exploitant le même système pour apprendre les modèles SLU pour la tâche visée, ce que nous faisons dans ce travail.

\begin{table}[th]
\begin{center}
\scalebox{0.6}{
    \begin{tabular}{l|c|c|c|c|c|c}

        \hline
        \multicolumn{7}{c}{Corpus: MEDIA, Métrique: taux d'erreur (CER)} \\
        \hline
        \textbf{Stg appr.} & \textbf{Entrée} & \textbf{KWh} (gCO$_2$) & \textbf{kWh/p} & \textbf{T appr.} & \textbf{DEV} & \textbf{TEST} \\
        \hline
        \hline
        \multicolumn{7}{|c|}{\textbf{Features de base}} \\
        \hline
        \cite{evain:hal-03407172} 3 steps & spectro & - & - & \textit{57h} & \textbf{29.07} & \textbf{31.10} \\
        3 steps   &   spectro & 6,651 (314) & 0,273 & 56h55' &   \textbf{28.35}  &   \textbf{28.95} \\
        2 steps   &   spectro & 4,417 (225) & 0.173 & 40h52' &   32.04  &   32.85 \\
        1 step   &   spectro & 2,407 (123) & $\mathcal{M}_e$ & 22h16' &   46.57  &   44.50 \\
        \hline
        \hline
        \cite{evain:hal-03407172} 3 steps & w2v2-fr & - & - & \textit{36h} & \textbf{17.25} & \textbf{16.25} \\
        3 steps &   w2v2-fr & 3.597 (183) & 0,550 & 36h01' &   \textbf{18.69}  &   \textbf{16.14} \\
        2 steps &   w2v2-fr & 2.445 (125) & 0,116 & 24h29' &   18.24  &   16.23 \\
        1 step  &   w2v2-fr & 2.150 (110) & $\mathcal{M}_e$ & 21h32' &   19.68  &   18.77 \\
        \hline
        \hline
        \multicolumn{7}{|c|}{\textbf{Features affinés (+100h x4 GPU)}} \\
        \hline
        2 steps +1   &   w2v2-fr slu & 2.569 (131) & $\infty$ & 27h28' &   14.25  &   13.78 \\
        1 step +1   &   w2v2-fr slu & 2.529 (129) & $\infty$ & 27h02' &   \textbf{14.16}  &   \textbf{13.26} \\
        \hline
        \cite{evain:hal-03407172}$^{(*)}$ 3 steps +1 & w2v2-fr asr & - & - & \textit{36h} & \textbf{14.58} & \textbf{13.78} \\
        \hline
        \hline
        \multicolumn{7}{|c|}{\textbf{Transfert}} \\
        \hline
        1 step +PM &   w2v2-fr & 2.420 (123) & 0,125 & 25h04' &   18.27  &   16.61 \\
        \hline
        \hline
        \multicolumn{7}{|c|}{\textbf{Transfert + features affinés (+100h x4 GPU)}} \\
        \hline
        1 step +1 +PM &   w2v2-fr slu & 2.026 (103) & $\mathcal{M}_e$ & 19h23' &   \textbf{13.59}  & \textbf{13.21} \\
        \hline
        \hline
        \multicolumn{7}{|c|}{\textbf{État de l'art}} \\
        \hline
        \cite{pelloin:hal-03128163} & MFCC & - & - & - & \textbf{16.1} & \textbf{13.6} \\
        \cite{ghannay:hal-03372494} & w2v2-fr slu$^{(**)}$ & - & - & - & - & \textbf{11.2} \\
        \hline
    \end{tabular}
}
\end{center}
\caption{Résultats sur le corpus MEDIA, pour tous les détails voir dans le texte. $^{(*)}$ le fine-tuning pour ces résultats était effectué pour l'ASR et non pas pour la SLU comme dans notre travail. $^{(**)}$ Features affinés en plusieurs étapes (ASR, SLU, modèle de langue) sur la tâche MEDIA.}
\label{tab:MEDIA-token-ICASSP2022}
\end{table}

\subsubsection{Experiences à moindre coût sur \emph{MEDIA}}

Les résultats sur la tâche MEDIA sont montrés dans le tableau~\ref{tab:MEDIA-token-ICASSP2022}. Dans le bloc \textbf{Features de base} sont montrés les résultats obtenus dans les mêmes conditions expérimentales que celles utilisées pour la tâche PortMEDIA. Ces résultats confirment qu'
un modèle SLU compétitif peut être obtenu à un moindre coût (\emph{3 steps} vs \emph{1 step} avec features \emph{w2v2-fr}).
Le bloc \textbf{Features affinés} montre les résultats obtenus avec des features extraits à partir du modèle SSL w2v2-fr affiné sur la tâche SLU de MEDIA (w2v2-fr slu).
Il est intéressant de noter que le modèle appris complètement de bout en bout (\emph{1 step +1}) obtient de meilleurs résultats que le modèle appris en 2 étapes (\emph{2 steps +1}).
Ceci grâce au fait que le modèle appris de bout en bout peut bénéficier d'un apprentissage plus \emph{agressif}, notamment une régularisation plus faible. Ces réglages ne sont pas efficaces avec une stratégie en 2 étapes, intuitivement parce qu’ils vont ``effacer'' l'information fournie par le modèle utilisé pour la préinitialisation, éloignant le modèle de l'optimum.
Puisque le modèle SSL est affiné sur la même tâche que le modèle SLU final, il n'est pas étonnant que ce dernier obtienne des résultats très compétitifs avec un coût d'apprentissage moindre. En effet, comparés aux derniers modèles à l'état de l'art sur la tâche MEDIA (bloc \textbf{État de l'art} dans le tableau), nos résultats sont meilleurs que ceux de \cite{pelloin:hal-03128163}, et assez proche de \cite{ghannay:hal-03372494}, alors qu'ils sont obtenus avec un coût bien moindre par rapport à ces travaux. \cite{pelloin:hal-03128163} et \cite{ghannay:hal-03372494} ne mentionne pas le coût computationnel de leur modèle, mais de ce qui est reporté dans leurs travaux il est possible d'estimer un coût supérieur à celui de l'affinage du modèle SSL que nous avons effectué (\cite{ghannay:hal-03372494} notamment effectue plusieurs de ces affinages). \\
Puisque les résultats commentés jusque là confirment que les modèles SSL permettent d'atteindre des performances très compétitives même avec un apprentissage de bout en bout (\emph{1 step}), pour les expériences suivantes nous avons utilisé uniquement cette stratégie d'apprentissage moins coûteuse. \\
Dans les blocs \textbf{Transfert} et \textbf{Transfert + features affinés} du tableau~\ref{tab:MEDIA-token-ICASSP2022} nous montrons respectivement les résultats obtenus avec apprentissage par transfert de la tâche PortMEDIA, et apprentissage par transfert de la même tâche en utilisant des features extraits avec le modèle SSL w2v2-fr affiné. 
Avec apprentissage par transfert seul (\textbf{Transfert}), utilisant un modèle entraîné sur PortMEDIA comme point de départ pour le modèle MEDIA (\textbf{+PM}), les résultats s'améliorent remarquablement sur les données de test (18.77 vs 16.61) sans aucun coût additionnel (en présumant qu'un modèle pour la tâche PortMEDIA soit disponible). La valeur \emph{kWh/p} des modèles utilisant les features \emph{w2v2-fr} est calculée par rapport à ce modèle (le plus économe avec cette entrée). \\
Avec apprentissage par transfert et features affinés pour la SLU (\textbf{Transfert + features affinés}), alors qu'il y a une amélioration sur les données de développement (14.16 vs 13.59), il n’y en a pratiquement pas sur les données de test (13.26 vs 13.21). Nous considérons que cela est dû au fait que le modèle SSL étant déjà affiné sur la tâche SLU, le petit apport des données PortMEDIA (ce corpus est même plus petit que MEDIA) à travers l'utilisation d'un modèle appris sur cette tâche n'ajoute pas plus d'information que celle déjà fournie par les features w2v2-fr affinées. Ce modèle a tout de même l'avantage d'être plus économe (19h23' vs 27h02'), toujours dans la perspective d'avoir un modèle SLU PortMEDIA déjà disponible en avance. La valeur \emph{kWh/p} pour les modèles utilisant des features \emph{w2v2-fr slu} est calculée par rapport à ce dernier modèle.


\section{Conclusions}

Dans cet article, nous avons analysé des stratégies d'apprentissage pour des modèles SLU sur la tâche MEDIA, visant à diminuer le coût computationnel de l'apprentissage des modèles tout en gardant des performances compétitives.
Nos résultats montrent que, en passant par l'utilisation de modèles SSL pour le français, il est possible d'atteindre ces objectifs.
Avec le coût additionnel de l'affinage d'un modèle SSL sur la tâche SLU, nous obtenons le deuxième meilleur résultat de la littérature sur MEDIA, et ce avec un entraînement complètement de bout en bout du modèle SLU. Bien que l'affinage soit relativement coûteux, notre modèle est bien moins gourmand en ressources que les meilleurs modèles de l'état de l'art.
Afin d'avoir une vision plus complète de l'impact énergique des modèles utilisés dans le TALN sur une tâche donnée, il serait souhaitable que la communauté adopte un standard d'évaluation des modèles d'un point de vue de leur consommation énergétique, notamment vis-à-vis de l'utilisation de plus en plus fréquente de modèles de plus en plus massifs et énergivores.


\section*{Remerciements}

Ce travail a été effectué en utilisant les ressources de calcul HPC de GENCI - IDRIS, numéro de contrat AD011011615R1. \\
Ce travail a été supporté partiellement par le projet JCJC CREMA (\emph{Coreference REsolution into MAchine translation}) financé par l'Agence Nationale de la Recherche (ANR), numéro de contrat ANR-21-CE23-0021-01.

\bibliographystyle{taln2022}
\bibliography{biblio}

\end{document}